\documentclass[10pt,twocolumn,letterpaper]{article}
\usepackage{cvpr}

\definecolor{cvprblue}{rgb}{0.21,0.49,0.74}
\usepackage[pagebackref,breaklinks,colorlinks,allcolors=cvprblue]{hyperref}

\usepackage{amsmath}

\usepackage{color}
\usepackage{multirow}
\usepackage{colortbl}
\usepackage{amssymb}

\usepackage{wrapfig}

\title{Learning Occlusion-Robust Vision Transformers for Real-Time UAV Tracking}

\author{
  You Wu$^{1\dagger}$,
  Xucheng Wang$^{2\dagger}$,
  Xiangyang Yang$^{1}$,
  Mengyuan Liu$^{1}$,\\
  Dan Zeng$^{3}$,
  Hengzhou Ye$^{1}$,
  Shuiwang Li$^{1*}$\\
  {$^{1}$College of Computer Science and Engineering, Guilin University of Technology, China}\\
  {$^{2}$School of Computer Science, Fudan University, Shanghai, China}\\
  {$^{3}$School of Artificial Intelligence, Sun Yat-sen University, Zhuhai, China}\\
  {\tt\small wuyou@glut.edu.cn, xcwang317@glut.edu.cn, xyyang317@163.com, mengyuaner1122@foxmail.com,}\\
  {\tt\small zengd8@mail.sysu.edu.cn, yehengzhou@glut.edu.cn, lishuiwang0721@163.com}
}

\begin{document}
\maketitle

\begin{abstract}
Single-stream architectures using Vision Transformer (ViT) backbones show great potential for real-time UAV tracking recently. However, frequent occlusions from obstacles like buildings and trees expose a major drawback: these models often lack strategies to handle occlusions effectively. New methods are needed to enhance the occlusion resilience of single-stream ViT models in aerial tracking.
In this work, we propose to learn Occlusion-Robust Representations (ORR) based on ViTs for UAV tracking by enforcing an invariance of the feature representation of a target with respect to random masking operations modeled by a spatial Cox process. Hopefully, this random masking approximately simulates target occlusions, thereby enabling us to learn ViTs that are robust to target occlusion for UAV tracking. This framework is termed ORTrack. Additionally, to facilitate real-time applications, we propose an Adaptive Feature-Based Knowledge Distillation (AFKD) method to create a more compact tracker, which adaptively mimics the behavior of the teacher model ORTrack according to the task's difficulty. This student model, dubbed ORTrack-D, retains much of ORTrack's performance while offering higher efficiency. Extensive experiments on multiple benchmarks validate the effectiveness of our method, demonstrating its state-of-the-art performance. Codes is available at \url{https://github.com/wuyou3474/ORTrack}.
\end{abstract}

\begin{figure}[t]
	\centering
\includegraphics[width=0.475\textwidth, height=0.29\textwidth]{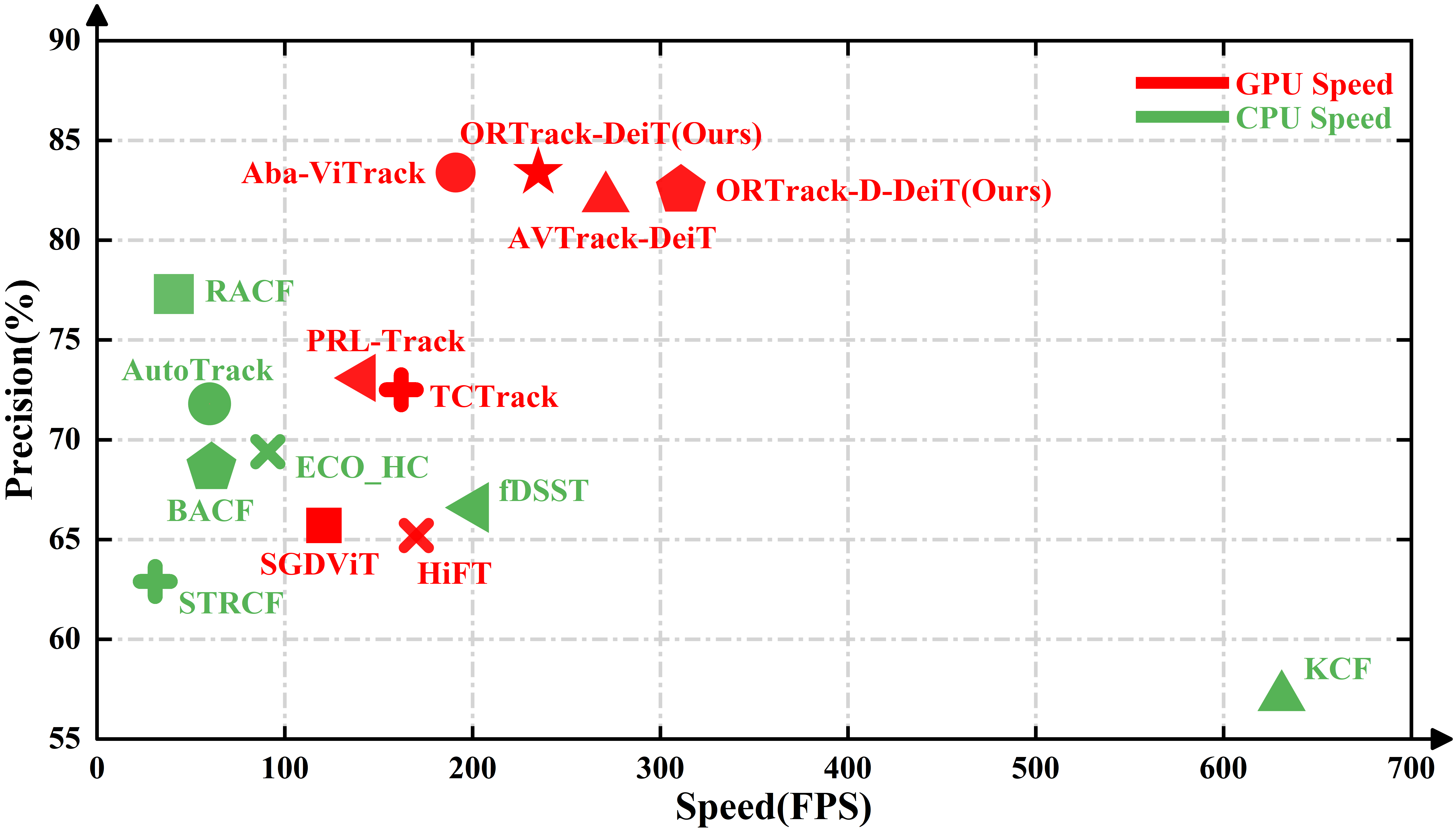}
\caption{Compared to SOTA UAV trackers on UAVDT, our ORTrack-DeiT sets a new record with 83.4\% precision and a speed of 236 FPS. Our ORTrack-D-DeiT strikes a better trade-off with 82.5\% precision and a speed of about 313 FPS.}
\label{fig_Prec_Speed}
\end{figure}

\section{Introduction}

\footnotetext[0]{$^\dagger$~Equal contribution. 
\quad $^*$~Corresponding authors.}

Unmanned aerial vehicles (UAVs) are leveraged in a plethora of applications, with increasing emphasis on UAV tracking \cite{li2020autotrack,cao2021hift,ma2023learning,Liu2022GlobalFP,li2024learning,wang2023learning,wu2024learning}. This form of tracking poses an exclusive set of challenges such as tricky viewing angles, motion blur, severe occlusions, and the need for efficiency due to UAVs' restricted battery life and computational resources \cite{Cao2022TCTrackTC,Wang2022RankBasedFP,Wu2022FisherPF,li2021learning}. 
Consequently, designing an effective UAV tracker requires a delicate balance between precision and efficiency. It needs to ensure accuracy while being conscious of the UAV's energy and computational constraints. 

In recent years, there has been a notable shift from discriminative correlation filters (DCF)-based methods, because of their unsatisfactory robustness, towards DL-based approaches, particularly with the adoption of single-stream architectures that integrate feature extraction and fusion via pre-trained Vision Transformer (ViT) backbone networks. This single-stream paradigm has proven highly effective in generic visual tracking, as evidenced by the success of recent methods such as OSTrack \cite{ye2022joint}, SimTrack \cite{chen2022backbone}, Mixformer \cite{cui2022mixformer}, and DropMAE \cite{Wu2023DropMAEMA}.
Building on these advancements, Aba-VTrack \cite{li2023adaptive} introduces a lightweight DL-based tracker within this framework, employing an adaptive and background-aware token computation method to enhance inference speed, which demonstrates remarkable precision and speed for real-time UAV tracking. However, the use of a variable number of tokens in Aba-VTrack incurs significant time costs, primarily due to the unstructured access operations required during inference. Adding to this, it also grappled with establishing robustness when facing target occlusion, a challenge common in UAV tracking often triggered by obstructive elements like buildings, mountains, trees, and so forth. 
The problem is exacerbated by the fact that UAVs may not always be capable of circumventing these impediments due to potential large-scale movements involved.

To address these issues, we introduce a novel framework designed to enhance the occlusion robustness of ViTs for UAV tracking. Our approach, termed ORTrack, aims to learn ViT-based trackers that maintain robust feature representations even in the presence of target occlusion. This is achieved by enforcing an invariance in the feature representation of the target with respect to random masking operations modeled by a spatial Cox process. The random masking serves as a simulation of target occlusion, which is expected to mimic real occlusion challenges in UAV tracking and aid in learning Occlusion-Robust Representations (ORR).
Notably, our method for learning occlusion-robust representation simply uses a Mean Squared Error (MSE) loss during training, adding no extra computational load during inference.
Additionally, to enhance efficiency for real-time applications, we introduce an Adaptive Feature-Based Knowledge Distillation (AFKD) method. This method creates a more compact tracker, named ORTrack-D, which adaptively mimics the behavior of the teacher model ORTrack based on the complexity of the tracking task during training. 
The reasoning is that the teacher model, in its pursuit of powerful representations, may compromise its generalizability. 
Hence, in situations where generalizability is vital, the student model may perform better, and closely mimicking the teacher's behavior becomes less important.
We use the deviation of GIoU loss \cite{Rezatofighi2019GeneralizedIO} from its average value to quantify the difficulty of the tracking task, which makes sense as loss value is a  commonly used criteria to define hard samples \cite{Shrivastava2016TrainingRO,Wang2018TowardsHC,Tu2023HierarchicallyCH}.
ORTrack-D maintains much of ORTrack's performance with higher efficiency, making it better suited for deployment in resource-constrained environments typical of UAV applications. 
Extensive experiments on four benchmarks show that our method achieves state-of-the-art performance.

In summary, our contributions are as follows: (i) We propose to learn Occlusion-Robust Representations (ORR) by imposing an invariance in the feature representation of the target with respect to random masking operations modeled by a spatial Cox process, which can be easily integrated into other tracking frameworks without requiring additional architectures or increasing inference time; (ii) We propose an Adaptive Feature-Based Knowledge Distillation (AFKD) method to further enhance efficiency, in which the student model adaptively mimics the behavior of the teacher model according to the task's difficulty, resulting in a significant increase in tracking speed while only minimally reducing accuracy; (iii) We introduce ORTrack, a family of efficient trackers based on these components, which integrates seamlessly with other ViT-based trackers. ORTrack demonstrates superior performance while maintaining extremely fast tracking speeds. Extensive evaluations show that ORTrack achieves state-of-the-art real-time performance.

\section{Related work}

\subsection{Visual Tracking. } 
In visual tracking, the primary approaches consist of DCF-based and DL-based trackers. DCF-based trackers are favored for UAV tracking due to their remarkable efficiency, but they face difficulties in maintaining robustness under complex conditions \cite{li2021learning,li2020autotrack,Huang2019LearningAR}. Recently developed lightweight DL-based trackers have improved tracking precision and robustness for UAV tracking \cite{cao2021hift,Cao2022TCTrackTC}; however, their efficiency lags behind that of most DCF-based trackers. Model compression techniques like those in \cite{Wang2022RankBasedFP,Wu2022FisherPF} have been used to further boost efficiency, yet these trackers still face issues with tracking precision.
Vision Transformers (ViTs) are gaining traction for streamlining and unifying frameworks in visual tracking, as seen in studies like \cite{Xie2021LearningTR, cui2022mixformer, ye2022joint, Xie2022CorrelationAwareDT,yang2025adaptively}. While these frameworks are compact and efficient, few are based on lightweight ViTs, making them impractical for real-time UAV tracking. To address this, Aba-ViTrack \cite{li2023adaptive} used lightweight ViTs and an adaptive, background-aware token computation method to enhance efficiency for real-time UAV tracking. However, the variable token number in this approach necessitates unstructured access operations, leading to significant time costs. In this work, we aim to improve the efficiency of ViTs for UAV tracking through knowledge distillation, a more structured method.

\subsection{Occlusion-Robust Feature Representation. } 
Occlusion-robust feature representation is crucial in computer vision and image processing. It involves developing methods that can recognize and process objects in images even when parts are hidden or occluded \cite{Wang2019RegionAN,Park2022HandOccNetO3}.
Early efforts often relied on handcrafted features, active appearance models, motion analysis, sensor fusion, etc \cite{Lowe1999ObjectRF,Storer2009ActiveAM,Irani1993MotionAF,Chang2001TrackingMP}. While effective in some cases, these methods struggled with the complexity and variability of real-world visual data.
The advent of deep learning revolutionized the field. Many studies have applied Convolutional Neural Networks (CNNs) and other deep architectures to extract occlusion-robust representations \cite{Wang2019RegionAN,Park2022HandOccNetO3,Qu2023TowardsNA,Jiang2024OcclusionrobustFR}. These approaches use deep models to capture complex patterns and variations in visual data, making learned features resilient to occlusions and having proven valuable for many computer vision applications, such as action recognition~\cite{Das2024OcclusionRS,Yang2021SelfSupervisedVP}, pose estimation~\cite{Park2022HandOccNetO3,Zhang20233DAwareNB}, and object detection~\cite{Chi2019PedHunterOR,Kim2020BBCNB}. The exploration of occlusion-robust representations in visual tracking has also demonstrated great success \cite{Nguyen2001OcclusionRA,Nguyen2004FastOO,Hariharakrishnan2005FastOT,Pan2007RobustOH,Zhang2012AKA,Israni2018FeatureDB, Kuipers2020HardOI,Askar2020OcclusionDA,Chakraborty2021LearningTT}. 
However, to our knowledge, there is a dearth of research to explore learning occlusion-robust ViTs particularly in a unified framework for UAV tracking.
In this study, we delve into the exploration of learning occlusion-robust feature representations based on ViTs by simulating occlusion challenges using random masking modeled by a spatial Cox process, specifically tailored for UAV tracking. This study represents the first use of ViTs for acquiring occlusion-robust feature representations in UAV tracking.

\subsection{Knowledge Distillation. }
Knowledge distillation is a technique used to compress models by transferring knowledge from a complex "teacher" model to a simpler "student" model, with the aim of maintaining performance while reducing computational resources and memory usage \cite{Park2019RelationalKD, Tung2019SimilarityPreservingKD}. It involves various types of knowledge, distillation strategies, and teacher-student architectures, typically falling into three categories: response-based, feature-based, and relation-based distillation \cite{gou2021knowledge, Wang2020KnowledgeDA, Park2019RelationalKD}. Widely applied in tasks such as image classification \cite{peng2019few}, object detection \cite{chen2017learning}, and neural machine translation \cite{li2021learning}, it offers potential to improve the efficiency and even effectiveness of deep learning models.
Recently, it has been successfully utilized to enhance the efficiency of DL-based trackers. For instance, Li et al. \cite{Li2022MaskGuidedSF} used mask-guided self-distillation to compress Siamese-based visual trackers. Sun et al. \cite{Sun2023SiamOHOTAL} introduced a lightweight dual Siamese tracker for hyperspectral object tracking, using a spatial-spectral knowledge distillation method to learn from a deep tracker. 
However, these techniques are mainly Siamese-based and tailored to specific tracking frameworks, posing challenges for adaptation to our ViT-based approach. In this study, we propose a simple yet effective feature-based knowledge distillation method, in which the student adaptively replicate the behavior of the teacher based on the complexity of the tracking task during training.

 \begin{figure*}[t]
\centering	\includegraphics[width=0.8\textwidth]{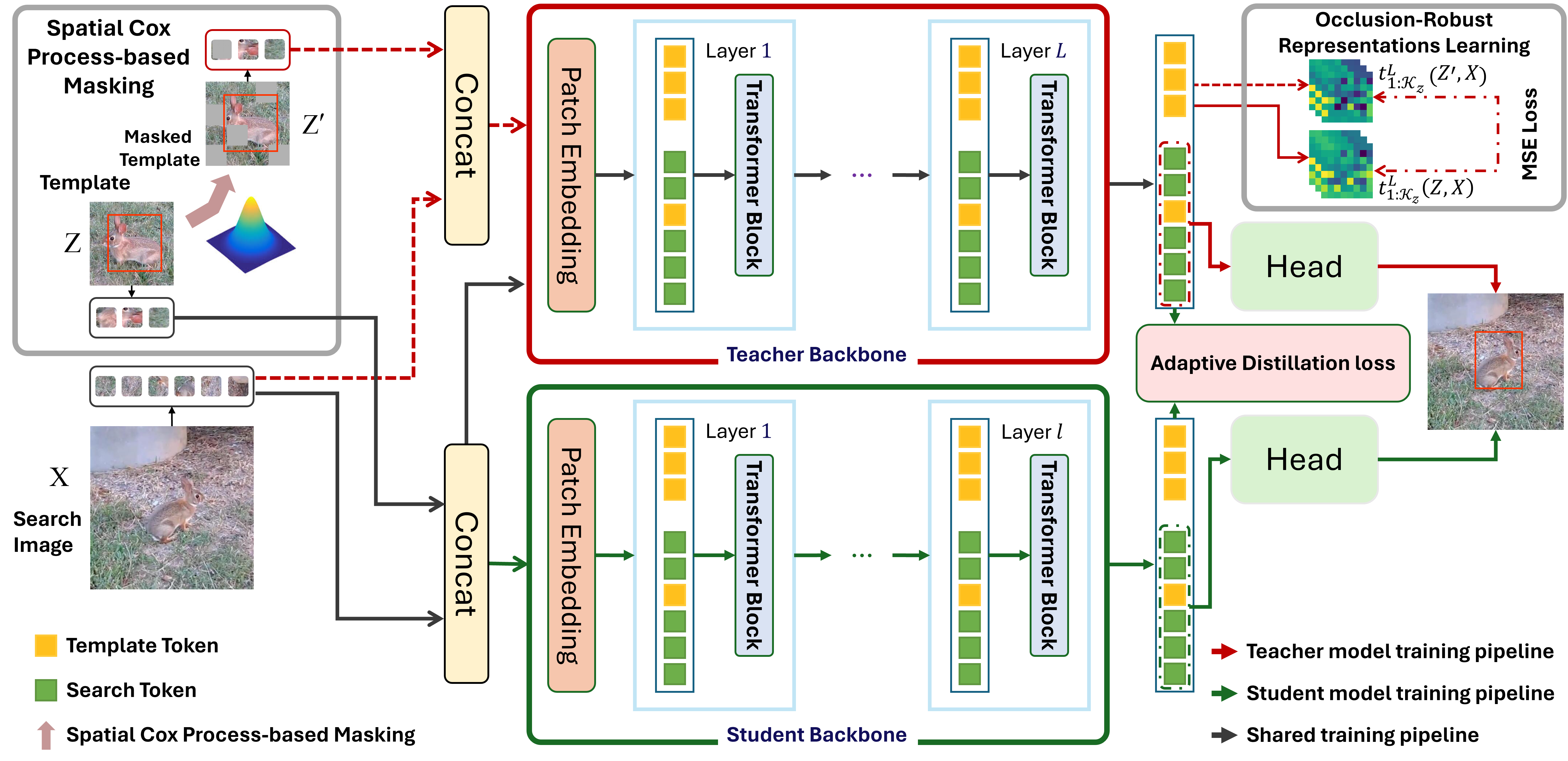}
\caption{Overview of the proposed ORTrack framework, which includes separate training pipelines for a teacher and a student model. Note that the spatial Cox process-based masking and occlusion-robust representation learning are applied only in the teacher pipeline. Once the teacher is trained, its weights are fixed for training the student model with the proposed adaptive knowledge distillation.} \label{fig:overview}
\end{figure*}

\section{Method}
	
 In this section, we first provide a brief overview of our end-to-end tracking framework, named ORTrack, as shown in Figure \ref{fig:overview}. Then, we introduce the occlusion-robust representation learning based on spatial Cox processes and the method of adaptive knowledge distillation. Finally, we detail the prediction head and training loss.

\subsection{Overview}

The proposed ORTrack introduces an novel single-stream tracking framework, featuring a spatial Cox process-based masking for occlusion-robust representation learning and an adaptive feature-based knowledge distillation pipeline. ORTrack consists of two sequential training phases: the teacher model training pipeline for learning occlusion-robust representations, followed by the student training pipeline involving adaptive knowledge distillation.
In the teacher model training phase, the input includes a target template $Z\in\mathbb{R}^{3\times H_{z}\times W_{z}}$ of spatial size $H_{z}\times W_{z}$, a randomly masked target template $Z'=\mathfrak{m}(Z)$, and a search image $X\in\mathbb{R}^{3\times H_{x}\times W_{x}}$ of spatial size $H_{x}\times W_{x}$, where $\mathfrak{m}(\cdot)$ represents the random masking operation that masks out non-overlap patches of size $b\times b$ with a certain masking ratio $\sigma$. 
To achieve occlusion-robust representation with ViTs, we minimize the mean squared error (MSE) between two versions of the template representation: one with random masking and one without. 
During the training of the student model, the teacher's weights remain fixed while both the teacher and student models receive inputs $Z$ and $X$. Let $\mathfrak{B}_T$ and $\mathfrak{B}_S$ represent the backbones of the teacher and student, respectively. 
In our implementation, $\mathfrak{B}_T$ and $\mathfrak{B}_S$ share the same structure of the ViT layer but differ in the number of layers. Feature-based knowledge distillation is used to transfer the knowledge embedded in the teacher model's backbone features to the student model through an adaptive distillation loss.

\subsection{Occlusion-Robust Representations (ORR) Based on Spatial Cox Processes}


To begin, we describe two random masking operations used to simulate occlusion challenges: one from MAE \cite{He2021MaskedAA} and our proposed method based on a Spatial Cox process, denoted by $\mathfrak{m}_{\textup{U}}$ and $\mathfrak{m}_{\textup{C}}$, respectively.
Although $\mathfrak{m}_{\textup{U}}$ allows the model to learn robust representations that are less sensitive to noise or missing information by randomly ignoring certain parts of the input data during training \cite{He2021MaskedAA}, it is less effective when used to simulate occlusion since each spatial position (in the sense of block size) is masked out with equal probability, especially in our situation where the target template generally contains background. 
To ensure that the target is masked out as expected with higher probabilities at a given masking ratio, thereby making the occlusion simulation more effective, we employ a finite Cox process \cite{Illian2008StatisticalAA} to model this masking operation, which is detailed as follows.

Define two associated random matrices $\mathbf{m}=(m_{i,j})$, $\mathbf{b}=(b_{i,j})$, ${1\leqslant i\leqslant H_z/b,1\leqslant j\leqslant W_z/b}$, where $m_{i,j}\backsim \mathcal{U}(0,1)$ (i.e., $m_{i,j}$ follows a uniform distribution over the interval [0, 1]), $b_{i,j}\in \{0,1\}$ equals $1$ if $m_{i,j}\in \textup{TopK}(\mathbf{m},K)$, and $0$ otherwise. $\textup{TopK}(\mathbf{m},K)$ returns the $K=\lfloor (1-\sigma)H_z W_z \rceil$ largest elements from $\mathbf{m}$, where 
$\lfloor x \rceil$ rounds $x$ to the nearest integer. Mathematically, $\mathfrak{m}_{\textup{U}}(Z)=Z\odot (\mathbf{b}\otimes \textbf{1})$, where $\odot$ denotes the Hadamard product and $\otimes$ denotes the tensor product,  $\textbf{1}$ is an all-ones matrix of size $b\times b$. Before defining $\mathfrak{m}_{\textup{C}}$, we establish core notations relevant to spatial Cox processes.
It extend the concept of spatial inhomogeneous Poisson point processes by incorporating a random intensity function, which, in turn, is defined as a Poisson point process with an intensity determined by a location-dependent function in the underlying space.
For Euclidean space $\mathbb{R}^2$, an inhomogeneous Poisson point process is defined by a locally integrable positive intensity function $\lambda\colon\mathbb{R}^2\to[0,\infty)$, such that for every bounded region $\mathcal{B}$ the integral $\Lambda (\mathcal{B})=\int_\mathcal{B} \lambda(x,y)\,\mathrm dxdy$ is finite, where $\Lambda (\mathcal{B})$ has the interpretation of being the expected number of points of the Poisson process located in $\mathcal{B}$, and for every collection of disjoint bounded Borel measurable sets $\mathcal{B}_1,...,\mathcal{B}_k$ \cite{ONeil2002GeometricMT}, its number distributions is defined by $\Pr \{\mathrm{N}(\mathcal{B}_i)=n_i,  i=1, \dots, k\}=\prod_{i=1}^k\frac{(\Lambda(\mathcal{B}_i))^{n_i}}{n_i!} e^{-\Lambda(\mathcal{B}_i)}$, $n_i\in \mathbb{Z}^{0+}$, 
where $\Pr$ denotes the probability measure, $\mathrm{N}$ indicates the random counting measure such that $\Lambda (\mathcal{B})= \mathbb{E}[\mathrm{N}(\mathcal{B})]$, $\mathbb{E}$ is the expectation operator. In particular,  the conditional distribution of the points in a bounded set $\mathcal{B}$ given that $\mathrm{N}(\mathcal{B}) = n \in \mathbb{Z}^{0+}$ is not uniform, and $f_{n}(p_1,...,p_n)=\prod_{n}^{i=1}\frac{\lambda(p_i) }{\Lambda (\mathcal{B})},\quad p_1,...,p_n\in \mathcal{B}$ defines the corresponding location density function of the $n$ points.
Since a Cox process can be regarded as the result of a two-stage random mechanism for which it is sometimes termed ‘doubly stochastic Poisson process’ \cite{Illian2008StatisticalAA}, the finite Cox processes can be simulated in a straightforward way based on
the hierarchical nature of the model. Specifically, in the first step, the intensity $\lambda(x,y)$ is generated. In the second step, an inhomogeneous Poisson point process is simulated using the generated  $\lambda(x,y)$ \cite{Illian2008StatisticalAA,Mattfeldt1996StochasticGA}.
The thinning algorithm \cite{Chen2016ThinningAF} is used here for simulating inhomogeneous Poisson point processes. It involves simulating a homogeneous Poisson point process with a higher rate than the maximum possible rate of the inhomogeneous process, and then "thinning" out the generated points to match the desired intensity function. 

In this work, the randomness of the intensity function is modeled by a random variable $\Gamma$ that has a Poisson distribution with expectation of $\varsigma$, namely, $\Pr\{\Gamma=k\}=\frac{\varsigma^ke^{-\varsigma}}{k!}$, where $k\in \mathbb{Z}^{0+}$. The intensity function of the inhomogeneous Poisson point process is then given by
\begin{equation}
\small
 \lambda(x,y)= \frac{\Gamma e^{-(x^2+y^2)}}{\int_{\mathcal{B}}e^{-(x^2+y^2) }dxdy}.
\end{equation}
Note that $\lambda(x,y)$ is a bell-shape function that gives more intensities to the central area of $\mathcal{B}$.
Let $\mathcal{B}$ denote the rectangle region of size $H_z/b\times W_z/b$ representing the template region. If we simulate the Cox process within $\mathcal{B}$ and denote a resulted point pattern by $\Xi$, we can obtain a matrix $\mathbf{b'}=(b'_{i,j})_{1\leqslant i\leqslant H_z/b,1\leqslant i\leqslant W_z/b}$, where $b'_{i,j}$ equals 1 if $(i,j)\in \Xi$, and 0 otherwise,
with which our $\mathfrak{m}_{\textup{C}}$ can be defined as $\mathfrak{m}_{\textup{C}}(Z)=Z\odot (\mathbf{b'}\otimes \textbf{1})$. It is worthy of note that if $\varsigma =\lfloor (1-\sigma)H_z W_z \rceil$, since $\mathbb{E}[\Lambda (\mathcal{B} )]=\mathbb{E}[\int_{\mathcal{B}}^{}\lambda (x,y)dxdy]=\mathbb{E}[\Gamma ]=\varsigma$,
in this case, the expected masking ratio of our masking operation is equal to the masking ratio of $\mathfrak{m}_{\textup{C}}$. 
Thus, in addition to inhomogeneous intensity, our method can simulate more diverse pattern of occlusion due to the introduced randomness of the masking ratio. 

We denote the total number of tokens by $\mathcal{K}$, the embedding dimension of each token by $d$, and all the tokens output by the $L$-th layer of $\mathfrak{B}_T$ with respect to inputs $X$ and $Z$ by $\mathbf{t}_{1:\mathcal{K}}^{L}(Z,X;\mathfrak{B}_T)\in \mathbb{R}^{\mathcal{K}\times d}$. 
Let $\mathbf{t}_{\mathcal{K}_Z\cup \mathcal{K}_X}^{L}(Z,X;\mathfrak{B}_T)=\mathbf{t}_{1:\mathcal{K}}^{L}(Z,X;\mathfrak{B}_T)$, where $\mathcal{K}_Z\cup\mathcal{K}_X=[1,\mathcal{K}]$,  $\mathbf{t}_{\mathcal{K}_Z}^{L}$ and $\mathbf{t}_{\mathcal{K}_X}^{L}$ represent the tokens corresponding to the template and the search image, respectively. By the same token, the output tokens corresponding to inputs $X$ and $Z'$ are $\mathbf{t}_{1:\mathcal{K}}^{L}(Z',X;\mathfrak{B}_T)$.
The feature representations of $Z$ and $Z'$ can be recovered by tracking their token indices in respective ordered sequences, which specifically are $t_{1:\mathcal{K}_z}^L(Z, X;\mathfrak{B}_T)$ and $t_{1:\mathcal{K}_z}^L(Z',X;\mathfrak{B}_T)$, respectively. The core idea of our occlusion-robust representations learning is that the mean square error between the feature representation of $Z$ and that of $Z'$ is minimized, which is implemented by minimizing the following MSE loss,
\begin{equation}\label{Eq_MI_loss}
		\small
\mathcal{L}_{orr}=||t_{1:\mathcal{K}_z}^L(Z,X;\mathfrak{B}_T)-t_{1:\mathcal{K}_z}^L(Z',X;\mathfrak{B}_T)||^2.
\end{equation}
During inference, only $[Z,X]$ is input to the model without the need for random template masking. Consequently, our method incurs no additional computational cost during inference. Notably, our method is independent of the ViTs used, any efficient ViTs can work in our framework. 

\subsection{Adaptive Feature-Based Knowledge Distillation (AFKD)}

Feature-based knowledge distillation is a technique in machine learning that trains a smaller student model to mimic a larger teacher model, which, instead of focusing only on final outputs, transfers intermediate features or representations from the teacher to the student \cite{gou2021knowledge, Wang2020KnowledgeDA}. This method uses the detailed internal representations from the teacher model to improve the student's learning process. However, there is a risk that the student model might overfit to the specific features of the teacher model, rather than generalizing well to new data. This can be particularly problematic if the teacher model has learned spurious correlations in the data. To combat this, we propose adaptively transferring knowledge based on the difficulty of the tracking task. We quantify this difficulty using the deviation of the GIoU loss \cite{Rezatofighi2019GeneralizedIO} (see Section \ref{subsection:3.4}) from its average value, calculated between the student's prediction and the ground truth.
Adapting knowledge transfer based on difficulty ensures that the student model doesn't heavily adjust its weights on easy tasks, which it can handle already probably due to its generalizability. Instead, it focuses more on challenging scenarios where its feature representation is less effective.

Additionally, the choice of teacher-student architectures is crucial in knowledge distillation. Given the wide array of possible student models, we adopt a self-similar approach where the student model mirrors the teacher's architecture but employs a smaller ViT backbone, using fewer ViT blocks. This strategy simplifies the design and eliminates the need for additional alignment techniques that would otherwise be necessary due to mismatched feature dimensions.
Lastly, layer selection and the metric of feature similarity are also crucial aspects of feature-based knowledge distillation. Given MSE's popularity in feature-based knowledge distillation and to avoid potential complexity associated with using multiple layers, we employ MSE to penalize differences between the output feature representations of both the teacher and student model's backbones, i.e., $t_{1:\mathcal{K}}^L(Z,X;\mathfrak{B}_T)$ and $t_{1:\mathcal{K}}^L(Z,X;\mathfrak{B}_S)$. The proposed adaptive knowledge distillation loss is defined by
\begin{equation}\label{Eq_MI_loss}
		\small
\mathcal{L}_{afkd}=(\alpha+\beta(\mathcal{L}_{iou}-\overline{\mathcal{L}_{iou}}))||t_{1:\mathcal{K}}^L(Z,X;\mathfrak{B}_T)-t_{1:\mathcal{K}}^L(Z,X;\mathfrak{B}_S)||^2,
\end{equation}
where $\alpha +\beta(\mathcal{L}_{iou}-\overline{\mathcal{L}_{iou}}):=\varpi (\mathcal{L}_{iou};\alpha,\beta)$ is a function of the deviation of GIoU loss from its average, with slop $\alpha $ and intercept $\beta$, used to quantify the difficulty of the tracking task.

\begin{table*}[t]
\scriptsize
\centering
\setlength\tabcolsep{4.0pt} 
\caption{Precision (Prec.), success rate (Succ.), and speed (FPS) comparison between ORTrack and		lightweight trackers on four UAV tracking benchmarks, i.e., DTB70 \cite{Li2017VisualOT},  UAVDT \cite{du2018the},  VisDrone2018 \cite{wen2018visdrone}, and UAV123 \cite{Mueller2016ABA}. {\color[HTML]{FE0000}Red}, {\color[HTML]{3531FF}blue} and {\color[HTML]{009901}green} indicate the first, second and third place. Note that the percent symbol (\%) is omitted for all Prec. and Succ. values.}
\label{tab:comparision_with_light_trackers}
\begin{tabular}{ccccccccccccccccc}
\toprule[1pt] 
\multicolumn{2}{c}{}                                                          &                                                         & \multicolumn{2}{c}{DTB70}                                                                                                   & \multicolumn{2}{c}{UAVDT}                                                                                                   & \multicolumn{2}{c}{VisDrone2018}                                                                                                & \multicolumn{2}{c}{UAV123}                                                                                                  & \multicolumn{2}{c}{Avg.}                                                                                                    & \multicolumn{2}{c}{Avg.FPS}                                                                                                  &                                                                          &                                                                         \\
\multicolumn{2}{c}{\multirow{-2}{*}{Method}}                                  & \multirow{-2}{*}{Source}                                & Prec.                                                        & Succ.                                                        & Prec.                                                        & Succ.                                                        & Prec.                                                        & Succ.                                                        & Prec.                                                        & Succ.                                                        & Prec.                                                        & Succ.                                                        & GPU                                                           & CPU                                                          & \multirow{-2}{*}{\begin{tabular}[c]{@{}c@{}}FLOPs\\ (GMac)\end{tabular}} & \multirow{-2}{*}{\begin{tabular}[c]{@{}c@{}}Param.\\  (M)\end{tabular}} \\ \hline
                            & KCF \cite{Henriques2014HighSpeedTW}                                              & TAPMI 15                                                & 46.8                                                         & 28.0                                                         & 57.1                                                         & 29.0                                                         & 68.5                                                         & 41.3                                                         & 52.3                                                         & 33.1                                                         & 56.2                                                         & 32.9                                                         & -                                                             & {\color[HTML]{FE0000} \textbf{624.3}}                        & -                                                                        & -                                                                       \\
                            & fDSST \cite{danelljan2017discriminative}                                           & TPAMI 17                                                & 53.4                                                         & 35.7                                                         & 66.6                                                         & 38.3                                                         & 69.8                                                         & 51.0                                                         & 58.3                                                         & 40.5                                                         & 62.0                                                         & 41.4                                                         & -                                                             & {\color[HTML]{3531FF} \textbf{193.4}}                        & -                                                                        & -                                                                       \\
                            & ECO\_HC \cite{Danelljan2016ECOEC}                                         & CVPR 17                                                 & 63.5                                                         & 44.8                                                         & 69.4                                                         & 41.6                                                         & 80.8                                                         & 58.1                                                         & 71.0                                                         & 49.6                                                         & 71.2                                                         & 48.5                                                         & -                                                             & {\color[HTML]{009901} \textbf{83.5}}                         & -                                                                        & -                                                                       \\
                            & AutoTrack \cite{li2020autotrack}                                        & CVPR 20                                                 & 71.6                                                         & 47.8                                                         & 71.8                                                         & 45.0                                                         & 78.8                                                         & 57.3                                                         & 68.9                                                         & 47.2                                                         & 72.8                                                         & 49.3                                                         & -                                                             & 57.8                                                         & -                                                                        & -                                                                       \\
\multirow{-5}{*}{\rotatebox{90}{DCF-based}} & RACF \cite{li2021learning}                                             & PR 22                                                   & 72.6                                                         & 50.5                                                         & 77.3                                                         & 49.4                                                         & 83.4                                                         & 60.0                                                         & 70.2                                                         & 47.7                                                         & 75.9                                                         & 51.8                                                         & -                                                             & 35.6                                                         & -                                                                        & -                                                                       \\ \hline
                            & HiFT \cite{cao2021hift}                                           & ICCV 21                                                 & 80.2                                                         & 59.4                                                         & 65.2                                                         & 47.5                                                         & 71.9                                                         & 52.6                                                         & 78.7                                                         & 59.0                                                         & 74.0                                                         & 54.6                                                         & 160.3                                                         & -                                                            & 7.2                                                                      & 9.9                                                                     \\
                            & TCTrack \cite{Cao2022TCTrackTC}                                          & CVPR 22                                                 & 81.2                                                         & 62.2                                                         & 72.5                                                         & 53.0                                                         & 79.9                                                         & 59.4                                                         & 80.0                                                         & 60.5                                                         & 78.4                                                         & 58.8                                                         & 149.6                                                         & -                                                            & 8.8                                                                      & 9.7                                                                     \\
                            & SGDViT \cite{yao2023sgdvit}                                          & ICRA 23                                                 & 78.5                                                         & 60.4                                                         & 65.7                                                         & 48.0                                                         & 72.1                                                         & 52.1                                                         & 75.4                                                         & 57.5                                                         & 72.9                                                         & 54.5                                                         & 110.5                                                         & -                                                            & 11.3                                                                     & 23.3                                                                    \\
             & DRCI \cite{zeng2023towards}                                           & ICME 23                                                 & 81.4                                                         & 61.8                                                         & {\color[HTML]{FE0000} \textbf{84.0}}                         & 59.0                                                         & 83.4                                                         & 60.0                                                         & 76.7                                                         & 59.7                                                         & 81.4                                                         & 60.1                                                         & {\color[HTML]{3531FF} \textbf{281.3}}                         & {\color[HTML]{3531FF} \textbf{62.7}}                         & 3.6                                                                      & 8.8                                                                     \\ 
\multirow{-5}{*}{\rotatebox{90}{CNN-based}} & PRL-Track \cite{fu2024progressive}                                            & IROS 24                                                 & 79.5                                                         & 60.6                                                         & 73.1                         & 53.5                                                         & 72.6                                                        & 53.8                                                        & 79.1                                                         & 59.3                                                         & 76.1                                                         & 56.8                                                         & 132.3                         & -                         & 7.4                                                                      & 12.0                                                                \\ \hline
                            & Aba-ViTrack \cite{li2023adaptive}                                    & ICCV 23                                                 & {\color[HTML]{3531FF} \textbf{85.9}}                         & {\color[HTML]{FE0000} \textbf{66.4}}                         & {\color[HTML]{3531FF} \textbf{83.4}}                         & {\color[HTML]{3531FF} \textbf{59.9}}                         & {\color[HTML]{3531FF} \textbf{86.1}}                         & {\color[HTML]{3531FF} \textbf{65.3}}                         & {\color[HTML]{FE0000} \textbf{86.4}}                         & {\color[HTML]{3531FF} \textbf{66.4}}                         & {\color[HTML]{3531FF} \textbf{85.5}}                         & {\color[HTML]{3531FF} \textbf{64.5}}                         & 181.5                                                         & 50.3                                                         & 2.4                                                                      & 8.0                                                                     \\
                            & SMAT \cite{gopal2024separable}                                           & WACV 24                                                 & 81.9                                                         & 63.8                                                         & 80.8                                                         & 58.7                                                         & 82.5                                                         & 63.4                                                         & 81.8                                                         & 64.6                                                         & 81.8                                                         & 62.6                                                         & 126.8                                                         & -                                                            & 3.2                                                                      & 8.6                                                                     \\
                            & AVTrack-DeiT \cite{lilearningicml}                                    & ICML 24                                                 & {\color[HTML]{009901} \textbf{84.3}}                         & {\color[HTML]{009901} \textbf{65.0}}                         & 82.1                                                         & 58.7                                                         & {\color[HTML]{009901} \textbf{86.0}}                         & {\color[HTML]{3531FF} \textbf{65.3}}                         & {\color[HTML]{3531FF} \textbf{84.8}}                         & {\color[HTML]{FE0000} \textbf{66.8}}                         & {\color[HTML]{009901} \textbf{84.2}}                         & {\color[HTML]{009901} \textbf{63.8}}                         & {\color[HTML]{009901} \textbf{260.3}}                         & {\color[HTML]{009901} \textbf{59.8}}                         & 0.97-1.9                                                                 & 3.5-7.9                                                                 \\
                            & \cellcolor[HTML]{F2F2FF}\textbf{ORTrack-DeiT}   & \cellcolor[HTML]{F2F2FF}                                & \cellcolor[HTML]{F2F2FF}{\color[HTML]{FE0000} \textbf{86.2}} & \cellcolor[HTML]{F2F2FF}{\color[HTML]{FE0000} \textbf{66.4}} & \cellcolor[HTML]{F2F2FF}{\color[HTML]{3531FF} \textbf{83.4}} & \cellcolor[HTML]{F2F2FF}{\color[HTML]{FE0000} \textbf{60.1}} & \cellcolor[HTML]{F2F2FF}{\color[HTML]{FE0000} \textbf{88.6}} & \cellcolor[HTML]{F2F2FF}{\color[HTML]{FE0000} \textbf{66.8}} & \cellcolor[HTML]{F2F2FF}{\color[HTML]{009901} \textbf{84.3}} & \cellcolor[HTML]{F2F2FF}{\color[HTML]{3531FF} \textbf{66.4}} & \cellcolor[HTML]{F2F2FF}{\color[HTML]{FE0000} \textbf{85.6}} & \cellcolor[HTML]{F2F2FF}{\color[HTML]{FE0000} \textbf{65.0}} & \cellcolor[HTML]{F2F2FF}226.4                                 & \cellcolor[HTML]{F2F2FF}55.4                                 & \cellcolor[HTML]{F2F2FF}2.4                                              & \cellcolor[HTML]{F2F2FF}7.9                                             \\
\multirow{-5}{*}{\rotatebox{90}{ViT-based}} & \cellcolor[HTML]{F2F2FF}\textbf{ORTrack-D-DeiT} & \multirow{-2}{*}{\cellcolor[HTML]{F2F2FF}\textbf{Ours}} 
& \cellcolor[HTML]{F2F2FF}83.7                                 & \cellcolor[HTML]{F2F2FF}{\color[HTML]{3531FF} \textbf{65.1}} & \cellcolor[HTML]{F2F2FF}{\color[HTML]{009901} \textbf{82.5}} & \cellcolor[HTML]{F2F2FF}{\color[HTML]{009901} \textbf{59.7}} & \cellcolor[HTML]{F2F2FF}84.6                                 & \cellcolor[HTML]{F2F2FF}{\color[HTML]{009901} \textbf{63.9}} & \cellcolor[HTML]{F2F2FF}84.0                                 & \cellcolor[HTML]{F2F2FF}{\color[HTML]{009901} \textbf{66.1}} & \cellcolor[HTML]{F2F2FF}83.7                                 & \cellcolor[HTML]{F2F2FF}63.7 & \cellcolor[HTML]{F2F2FF}{\color[HTML]{FE0000} \textbf{292.3}} & \cellcolor[HTML]{F2F2FF}{\color[HTML]{FE0000} \textbf{64.7}} & \cellcolor[HTML]{F2F2FF}1.5                                              & \cellcolor[HTML]{F2F2FF}5.3           \\ \bottomrule[1pt]                                 
\end{tabular}
\end{table*}

\subsection{Prediction Head and Training Loss}
\label{subsection:3.4}

Following the corner detection head in \cite{cui2022mixformer,ye2022joint}, we use a prediction head consisting of multiple Conv-BN-ReLU layers to directly estimate the bounding box of the target. The output tokens corresponding to the search image are first reinterpreted to a 2D spatial feature map and then fed into the prediction head. 
The head outputs a local offset $\mathbf{o}\in [0,1]^{2\times H_x/P\times W_x/P}$, a normalized bounding box size $\mathbf{s} \in [0,1]^{2\times H_x/P\times W_x/P}$, and a target classification score $\mathbf{p} \in [0,1]^{H_x/P\times W_x/P}$ as prediction outcomes.
The initial estimation of the target position depends on identifying the location with the highest classification score, i.e., $(x_c, y_c)=\textup{argmax}_{(x,y)}\mathbf{p}(x,y)$. The final target bounding box is estimated by $\{(x_t,y_t);(w,h)\}=\{(x_c, y_c)+\mathbf{o}(x_c,y_c);\mathbf{s}(x_c,y_c)\}$.
For the tracking task, we adopt the weighted focal loss \cite{Law2018CornerNetDO} for classification, a combination of $L_1$ loss and GIoU loss \cite{Rezatofighi2019GeneralizedIO} for bounding box regression. The total loss for tracking prediction is:
\begin{equation}
\small
    \label{eq_loss}
    \mathcal{L}_{pred} =  \mathcal{L}_{cls} + \lambda_{iou}\mathcal{L}_{iou} + \lambda_{L_{1}}\mathcal{L}_{L_{1}},
\end{equation}
where the constants $\lambda_{iou}$ = 2 and $\lambda_{L_{1}}$= 5 are set as in \cite{cui2022mixformer,ye2022joint}.
The overall loss $\mathcal{L}_{T} =  \mathcal{L}_{pred} + \gamma\mathcal{L}_{orr}$ is used to train the teacher end-to-end after loading the pretrained weights of the ViT trained with ImageNet \cite{Russakovsky2014ImageNetLS}, where the constant $\gamma$ is set to $2.0 \times 10^{-4}$. After this training, we fix the weights of the teacher model, and employ the overall loss $\mathcal{L}_{S} =  \mathcal{L}_{pred} + \mathcal{L}_{afkd},$ for end-to-end knowledge distillation training.

\section{Experiments}

We evaluate our method on four UAV tracking benchmarks: DTB70 \cite{Li2017VisualOT}, UAVDT \cite{du2018the}, VisDrone2018 \cite{wen2018visdrone}, and UAV123 \cite{Mueller2016ABA}. All experiments run on a PC with an i9-10850K processor, 16GB RAM, and an NVIDIA TitanX GPU. We compare our method against 26 state-of-the-art trackers, using their official codes and hyper-parameters. 
We evaluate our approach against 13 state-of-the-art (SOTA) lightweight trackers (see Table \ref{tab:comparision_with_light_trackers}) and 14 SOTA deep trackers designed specifically for generic visual tracking (refer to Table \ref{tab:comparision_with_deep_trackers}).

\begin{table*}[t]
\scriptsize
\centering
\setlength\tabcolsep{4.0pt} 
\caption{Precision (Prec.) and speed (FPS) comparison between ORTrack-DeiT and deep-based trackers on VisDrone2018 \cite{wen2018visdrone}. }
\label{tab:comparision_with_deep_trackers}
\begin{tabular}{c|c|ccc|c|c|ccc|c|c|ccc}
\toprule[1pt]
Tracker                                       & Source                                & Prec.                                                        & Succ.                                                        & FPS                                                           & Tracker   & Source   & Prec. & Succ. & FPS                                  & Tracker   & Source  & Prec. & Succ. & FPS                                  \\
\hline
\cellcolor[HTML]{F2F2FF}\textbf{ORTrack-DeiT} & \cellcolor[HTML]{F2F2FF}\textbf{Ours} & \cellcolor[HTML]{F2F2FF}{\color[HTML]{FE0000} \textbf{88.6}} & \cellcolor[HTML]{F2F2FF}{\color[HTML]{009901} \textbf{66.8}} & \cellcolor[HTML]{F2F2FF}{\color[HTML]{FE0000} \textbf{206.2}} & ZoomTrack \cite{kou2023zoomtrack} & NIPS 23  & 81.4  & 63.4  & 61.7                                 & SimTrack \cite{chen2022backbone} & ECCV 22 & 80.0  & 60.9  & {\color[HTML]{3531FF} \textbf{69.7}} \\
AQATrack \cite{xie2024autoregressive}                                      & CVPR 24                               & {\color[HTML]{3531FF} \textbf{87.2}}                         & {\color[HTML]{3531FF} \textbf{66.9}}                         & 53.4                                                          & SeqTrack \cite{Chen2023SeqTrackST}  & CVPR 23  & 85.3  & 65.8  & 15.3                                 & ToMP \cite{Mayer2022TransformingMP}      & CVPR 22 & 84.1  & 64.4  & 21.4                                 \\
HIPTrack \cite{cai2024hiptrack}                                      & CVPR 24                               & {\color[HTML]{009901} \textbf{86.7}}                         & {\color[HTML]{FE0000} \textbf{67.1}}                         & 31.3                                                          & MAT \cite{zhao2023representation}       & CVPR 23  & 81.6  & 62.2  & {\color[HTML]{009901} \textbf{68.4}} & KeepTrack \cite{Mayer2021LearningTC} & ICCV 21 & 84.0  & 63.5  & 20.3                                 \\
EVPTrack \cite{shi2024evptrack}                                       & AAAI 24                               & 84.5                                                         & 65.8                                                         & 22.1                                                          & SparseTT \cite{2022SparseTT} & IJCAI 22 & 81.4  & 62.1  & 30.2                                 & SAOT \cite{Zhou2021SaliencyAssociatedOT}      & ICCV 21 & 76.9  & 59.1  & 35.4                                 \\
ROMTrack \cite{Cai_2023_ICCV}                                      & ICCV 23                               & 86.4                                                         & 66.7                                                         & 51.1                                                          & OSTrack \cite{ye2022joint}    & ECCV 22  & 84.2  & 64.8  & 62.7                                 & PrDiMP50 \cite{Danelljan2020ProbabilisticRF}  & CVPR 20 & 79.4  & 59.7  & 42.6 \\ \bottomrule[1pt]                               
\end{tabular}
\end{table*}

\subsection{Implementation Details}
  We adopt different ViTs as backbones, including ViT-tiny \cite{Dosovitskiy2020AnII}, Eva-tiny \cite{fang2023eva}, and DeiT-tiny \cite{touvron2021training}, to build three trackers for evaluation: ORTrack-ViT, ORTrack-Eva, and ORTrack-DeiT. 
  The head of ORTrack consists of a stack of four Conv-BN-ReLU layers. The search region and template sizes are set to 256 × 256 and 128 × 128, respectively.
A combination of training sets from GOT-10k \cite{2021GOT}, LaSOT \cite{Fan2018LaSOTAH}, COCO \cite{2014Microsoft}, and TrackingNet \cite{2018TrackingNet} is used for the training. The batch size is set to 32. We employ the AdamW optimizer \cite{Loshchilov2017DecoupledWD}, with a weight decay of $10^{-4}$ and an initial learning rate of $4 \times 10^{-5}$. 
The training is conducted over 300 epochs, with 60,000 image pairs processed in each epoch. The learning rate is reduced by a factor of 10 after 240 epochs. 

\subsection{State-of-the-art Comparison}



\textbf{Comparison with Lightweight Trackers. }
The overall performance of our ORTrack in comparison to 13 competing trackers on the four benchmarks is displayed in Table \ref{tab:comparision_with_light_trackers}.
As can be seen, our trackers demonstrate superior performance among all these trackers in terms of average (Avg.) precision (Prec.), success rate (Succ.) and speeds.
On average, RACF \cite{li2021learning} demonstrated the highest Prec. (75.9\%) and Succ. (51.8\%) among DCF-based trackers, DRCI \cite{zeng2023towards} achieves the highest precision and success rates, with 81.4\% and 60.1\%, respectively, among CNN-based trackers.
However, the average Prec. and Succ. of all our trackers are greater than 82.0\% and 62.0\%, respectively, clearly surpassing DCF- and CNN- based approaches. Additionally, our ORTrack-DeiT achieves the highest Avg. Prec. and Avg. Succ. of 85.6\% and 65.0\%, respectively, among all competing trackers.
Although Aba-ViTrack achieves performance close to our ORTrack-DeiT, its GPU speed is significantly lower, with a 23.6\%  relative gap.
Notably, when the proposed adaptive knowledge distillation is applied to ORTrack-DeiT, the resulting student model, ORTrack-D-DeiT, shows a significant speed increase: 29.1\% on GPU and 16.8\% on CPU. 
This improvement is accompanied by a minimal reduction in accuracy, with only a 1.9\% decrease in Avg. Prec. and a 1.3\% decrease in Avg. Succ..
All proposed trackers can run in real-time on a single CPU\footnote{Real-time performance applies to platforms similar to or more advanced than ours.}, and our ORTrack-DeiT sets a new performance record for real-time UAV tracking. 
We also compare the floating point operations per second (FLOPs) and number of parameters (Params.) of our method with CNN-based and ViT-based trackers in Table \ref{tab:comparision_with_light_trackers}.
Our method demonstrates a relatively lower parameter count and reduced computational complexity compared to these approaches.
Notably, since AVTrack-DeiT tracker features adaptive architectures, the FLOPs and parameters range from minimum to maximum values.
These results highlight our method's effectiveness and its state-of-the-art performance.

\textbf{Comparison with Deep Trackers. }
The proposed ORTrack-DeiT is also compared with 14 SOTA deep trackers in Table \ref{tab:comparision_with_deep_trackers}, which shows precision (Prec.) and GPU speed on VisDrone2018. Our ORTrack-DeiT surpasses all other methods in both metrics, demonstrating its superior accuracy and speed. Although trackers like AQATrack \cite{xie2024autoregressive}, HIPTrack \cite{cai2024hiptrack}, and ROMTrack \cite{Cai_2023_ICCV} achieve precision  comparable to our ORTrack-DeiT, their GPU speeds are much slower. Specifically, our method is 4, 6, and 4 times faster than AQATrack, HIPTrack, and ROMTrack, respectively.

\begin{figure}[h]
	\centering
\includegraphics[width=0.475\textwidth]{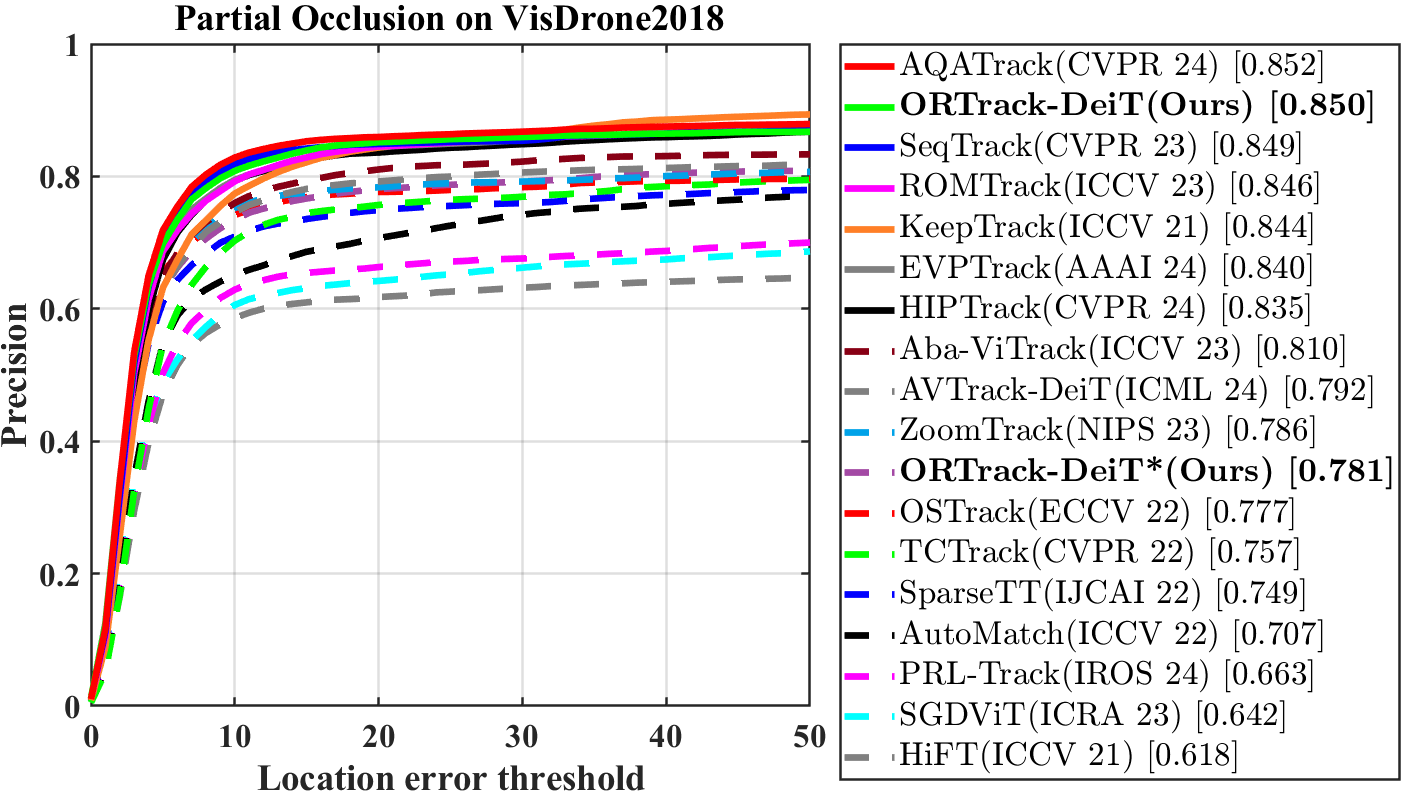}
\caption{Attribute-based comparison on the partial occlusion subset of VisDrone2018 \cite{wen2018visdrone}. ORTrack-DeiT* refers to ORTrack-DeiT without applying the occlusion-robust enhancement.} \label{fig_visdrone_attr}
\end{figure}

\textbf{Attribute-Based Evaluation. }
To access our method's robustness against target occlusion, we compare ORTrack-DeiT alongside 16 SOTA trackers on the partial occlusion subset of VisDrone2018. 
Additionally, we also assess the baseline ORTrack-DeiT*, i.e.,  ORTrack-DeiT without applying the proposed method for learning Occlusion-Robust Representation (ORR), for comparison.
The precision plot are presented in Fig. \ref{fig_visdrone_attr}, with additional attribute-based evaluation results provided in the supplemental materials. As observed, ORTrack-DeiT achieves the second-highest precision (85.0\%), just slightly behind the first-ranked tracker AQATrack by 0.2\%. Remarkably, incorporating the proposed components leads to a significant improvement over ORTrack-DeiT*, with increases of 6.9\% in Prec., well underscoring the effectiveness of our method.

\subsection{Ablation Study}

\begin{table}[h]
\scriptsize
\centering
\setlength\tabcolsep{4.0pt} 
\caption{Effect of ORR and AFKD on the baseline trackers.}
\label{tab:differentBackbone1}
\begin{tabular}{cccccc}
\toprule[1pt]      
\multirow{2}{*}{Tracker}      & \multirow{2}{*}{ORR} & \multirow{2}{*}{AFKD} & \multicolumn{2}{c}{UAVDT}     & \multirow{2}{*}{FPS} \\
                              &                      &                       & Prec.         & Succ.         &                      \\ \hline
\multirow{3}{*}{ORTrack-ViT}  &                      &                       & 77.0          & 55.6          & 216.2                \\
                              &$\checkmark$                       &                       & \textbf{80.3}$_{\uparrow 3.3}$ & \textbf{58.2}$_{\uparrow 2.6}$ & -                    \\
                              &$\checkmark$                       &$\checkmark$                        & 79.1$_{\uparrow 2.1}$          & 57.5$_{\uparrow 1.9}$          & \textbf{290.3}$_{\uparrow 34\%}$       \\ \hline
\multirow{3}{*}{ORTrack-Eva}  &                      &                       & 78.1          & 56.6          & 238.3                \\
                              &$\checkmark$                      &                       & \textbf{80.8}$_{\uparrow 2.7}$ & \textbf{58.7}$_{\uparrow 2.1}$ & -                    \\
                              &$\checkmark$                      &$\checkmark$                       & 79.5$_{\uparrow 1.4}$          & 57.8$_{\uparrow 1.2}$          & \textbf{308.8}$_{\uparrow 30\%}$       \\ \hline
\multirow{3}{*}{ORTrack-DeiT} &                      &                       & 78.6          & 56.7          & 218.4                \\
                              & $\checkmark$                     &                       & \textbf{83.4}$_{\uparrow 4.8}$ & \textbf{60.1}$_{\uparrow 3.4}$ & -                    \\
                              &$\checkmark$                      &$\checkmark$                       & 82.5$_{\uparrow 3.9}$          & 59.7$_{\uparrow 3.0}$          & \textbf{298.7}$_{\uparrow 36\%}$      \\   \bottomrule[1pt] 
\end{tabular}
\end{table}

\textbf{Effect of Occlusion-Robust Representations (ORR) and Adaptive Feature-Based Knowledge Distillation (AFKD). }
To demonstrate the effectiveness of the proposed ORR and AFKD, Table \ref{tab:differentBackbone1} shows the evaluation results on UAVDT dataset as these components are gradually integrated into the baselines. To avoid potential variations due to randomness, we only present the speed of the baseline, since the GPU speeds of the baseline and its ORR-enhanced version are theoretically identical.
As can bee seen, the incorporation of ORR significantly enhances both Prec. and Succ. for all baseline trackers. Specifically, the Prec. increases for ORTrack-ViT, ORTrack-Eva, and ORTrack-DeiT are 3.3\%, 2.7\%, and 4.8\%, respectively, while the Succ. increases are 2.6\%, 2.1\%, and 3.1\%, respectively. These significant enhancements highlight the effectiveness of ORR in improving tracking precision.
The further integration of AFKD results in consistent improvements in GPU speeds, with only slight reductions in Prec. and Succ. Specifically, all baseline trackers experience GPU speed enhancements of over 30.0\%, with ORTrack-DeiT showing an impressive 36.0\% improvement. These results affirm the effectiveness of AFKD in optimizing tracking efficiency while maintaining high tracking performance.


\begin{table}[h]
\centering
\scriptsize
\setlength\tabcolsep{1.5pt}
\caption{Impact of various Masking Operators on performance.}
\label{tab_effect_of_MO}
\begin{tabular}{cccccc|cc}
\toprule[1pt]
\multirow{2}{*}{Method}        & \multirow{2}{*}{$\mathfrak{m}_{\textup{U}}$} & \multirow{2}{*}{$\mathfrak{m}_{\textup{C}}$} & \multirow{2}{*}{SAM\cite{kirillov2023segment}} & \multirow{2}{*}{AdAutoMix\cite{qin2023adversarial}} & \multirow{2}{*}{CutMix\cite{yun2019cutmix}} & \multicolumn{2}{c}{VisDrone2018} \\
                               &                     &                     &                         &                        &                      & Prec.           & Succ.          \\ \hline
\multirow{6}{*}{ORTrack-DeiT} &                     &                     &                         &                        &                      & 81.6            & 62.2           \\
                               & $\checkmark$                   &                     &                         &                        &                      & 86.7            & 65.4           \\ 
                               &                     & $\checkmark$                   &                         &                        &                      & \textbf{88.6}   & \textbf{66.8}  \\
                               &                     &                     & $\checkmark$                       &                        &                      & 86.8            & 65.6           \\
                               &                     &                     &                         & $\checkmark$                      &                      & 84.3            & 63.8           \\
                               &                     &                     &                         &                        & $\checkmark$                    & 85.7            & 64.2    \\ \bottomrule[1pt]     
\end{tabular}
\end{table}

\textbf{Effect of Masking Operators. } 
To demonstrate the superiority of the proposed masking operator in terms of performance, we evaluate ORTrack-DeiT with various implementations of masking operators (i.e., $\mathfrak{m}_{\textup{U}}$, $\mathfrak{m}_{\textup{C}}$, and SAM \cite{kirillov2023segment}) alongside data mixing augmentation methods (i.e., AdAutoMix \cite{qin2023adversarial} and CutMix \cite{yun2019cutmix}).
The evaluation results on VisDrone2018 are presented in Table \ref{tab_effect_of_MO}. 
As shown, although using SAM, AdAutoMix, and CutMix improves performance, the best result achieved with SAM is only comparable to the performance of our $\mathfrak{m}_{\textup{U}}$ masking operator.
When $\mathfrak{m}_{\textup{C}}$ is applied, the improvements are even more substantial, with increases of 7.0\% and 4.6\%, respectively. These results validate the effectiveness of the proposed ORR component and particularly demonstrate the superiority of the masking operator based on spatial Cox processes.

\begin{table}[h]
        \centering
       \scriptsize
\setlength\tabcolsep{2.5pt} 
\caption{Impact of the adaptive knowledge distillation loss on the generalizability on LaSOT and TrackingNet.}
\begin{tabular}{cccccc|ccc}
        \toprule[1pt] 
\multirow{2}{*}{Method}      & \multirow{2}{*}{KD} & \multirow{2}{*}{AFKD} & \multicolumn{3}{c|}{LaSOT} & \multicolumn{3}{c}{TrackingNet} \\ 
                              &                     &                       & AUC    & Pnorm   & P      & AUC      & Pnorm     & P        \\ \hline
\multirow{3}{*}{ORTrack-DeiT} &                     &                       & 53.7   & 60.8    & 52.6   & 72.8     & 77.8      & 67.1     \\ 
                              &$\checkmark$                     &                       & 54.0   & 61.2   & 53.2   & 73.1     & 78.4      & 67.4     \\
                              &                     &  $\checkmark$                      & \textbf{54.6}   & \textbf{62.6}    & \textbf{54.3}   & \textbf{73.7}     & \textbf{79.1}      & \textbf{68.2}    \\         \bottomrule[1pt] 
\end{tabular}\label{tab_impact_akd}
\end{table}

\textbf{Impact of the Adaptive Knowledge Distillation Loss. } To assess the impact of the adaptive knowledge distillation loss on generalizability, we train ORTrack-DeiT using GOT-10K with 
$\varpi (\mathcal{L}_{iou};\alpha,\beta)$ and $\varpi (\mathcal{L}_{iou};\alpha, 0)$ separately, then evaluate them on LaSOT and TrackingNet. The results are shown in Table \ref{tab_impact_akd}. 
Note that $\varpi (\mathcal{L}_{iou};\alpha, 0)$ degenerates to a non-adaptive knowledge distillation loss as it becomes a constant. As can be seen, AFKD demonstrates greater performance improvements than KD. For instance, using AFKD results in additional gains of over 1.1\% in $P_{norm}$ and $P$ on LaSOT, demonstrating its superior generalizability.

\begin{table}[h]
\centering
\scriptsize
\setlength\tabcolsep{4.5pt} 
\caption{Application of our ORR component to three SOTA trackers: ARTrack \cite{wei2023autoregressive}, GRM \cite{gao2023generalized}, and DropTrack\cite{Wu2023DropMAEMA}.}
\label{table_sota}
\begin{tabular}{cccccc}
\toprule[1pt]
\multirow{2}{*}{Tracker}   & \multirow{2}{*}{ORR} & \multicolumn{2}{c}{UAVDT}     & \multicolumn{2}{c}{VisDrone2018} \\
                           &                      & Prec.         & Succ.         & Prec.           & Succ.          \\ \hline
\multirow{2}{*}{ARTrack\cite{wei2023autoregressive}}   &                      & 77.1          & 54.6          & 77.7            & 59.5           \\
                           & $\checkmark$                       & \textbf{78.5}$_{\uparrow 1.4}$ & \textbf{55.8}$_{\uparrow 1.2}$ & \textbf{79.5}$_{\uparrow 1.8}$   & \textbf{60.8}$_{\uparrow 1.3}$  \\ \hline
\multirow{2}{*}{GRM\cite{gao2023generalized}}       &                      & 79.0          & 57.7          & 82.7            & 63.4           \\
                           &  $\checkmark$                      & \textbf{81.7}$_{\uparrow 1.7}$ & \textbf{59.3}$_{\uparrow 1.6}$ & \textbf{84.8}$_{\uparrow 2.1}$   & \textbf{64.6}$_{\uparrow 1.2}$  \\ \hline
\multirow{2}{*}{DropTrack\cite{Wu2023DropMAEMA}} &                      & 76.9          & 55.9          & 81.5            & 62.7           \\
                           & $\checkmark$                       & \textbf{78.7}$_{\uparrow 1.8}$ & \textbf{57.4}$_{\uparrow 1.5}$ & \textbf{82.8}$_{\uparrow 1.3}$   & \textbf{64.2}$_{\uparrow 1.5}$  \\ \bottomrule[1pt]
\end{tabular}
\end{table}

\textbf{Application to SOTA trackers.}
To show the wide applicability of our proposed method, we incorporate the proposed ORR into three existing SOTA trackers: ARTrack \cite{wei2023autoregressive}, GRM \cite{gao2023generalized}, and DropTrack \cite{Wu2023DropMAEMA}.
Please note that we replace the model's original backbones with ViT-tiny \cite{Dosovitskiy2020AnII} to reduce training time.
As shown in Table \ref{table_sota}, incorporating ORR results in significant improvements in both precision and success rates for the three baseline trackers. 
Specifically, ARTrack, GRM, and DropTrack demonstrate an improvement of more than 1.2\% in both precision and success rate across two datasets.
These experimental results demonstrate that the proposed ORR component can be seamlessly integrated into existing tracking frameworks, significantly improving tracking accuracy.

\begin{figure}[h]
\centering
\includegraphics[width=0.475\textwidth]{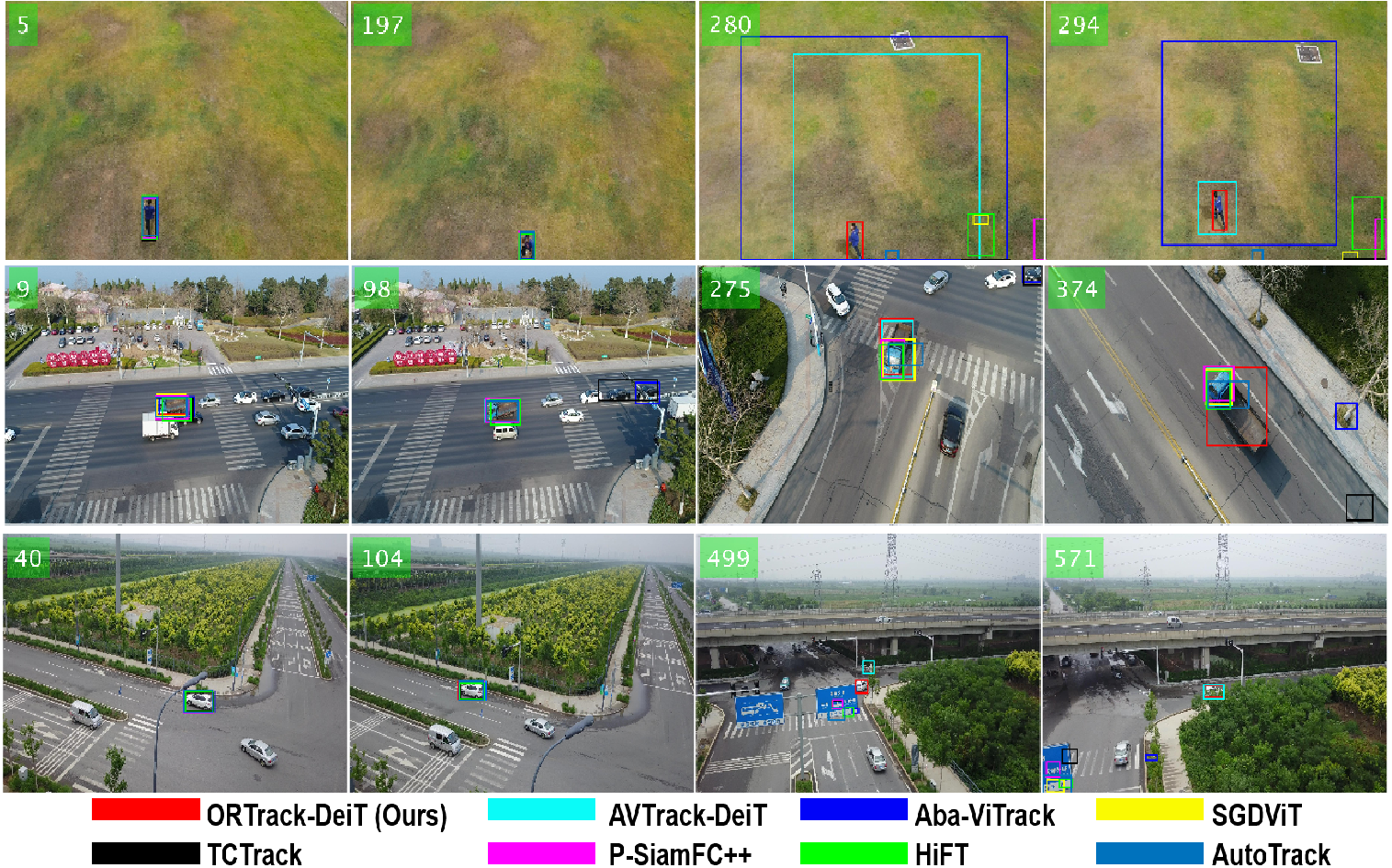}
\caption{Qualitative evaluation on 3 video sequences from, respectively, UAV123 \cite{Mueller2016ABA}, UAVDT \cite{du2018the}, and VisDrone2018 \cite{wen2018visdrone} (i.e., person9, S1607, and uav0000180\_00050\_s).} \label{fig_bbox_vis}
\end{figure}


\begin{figure}[h]
\centering
\includegraphics[width=0.475\textwidth]{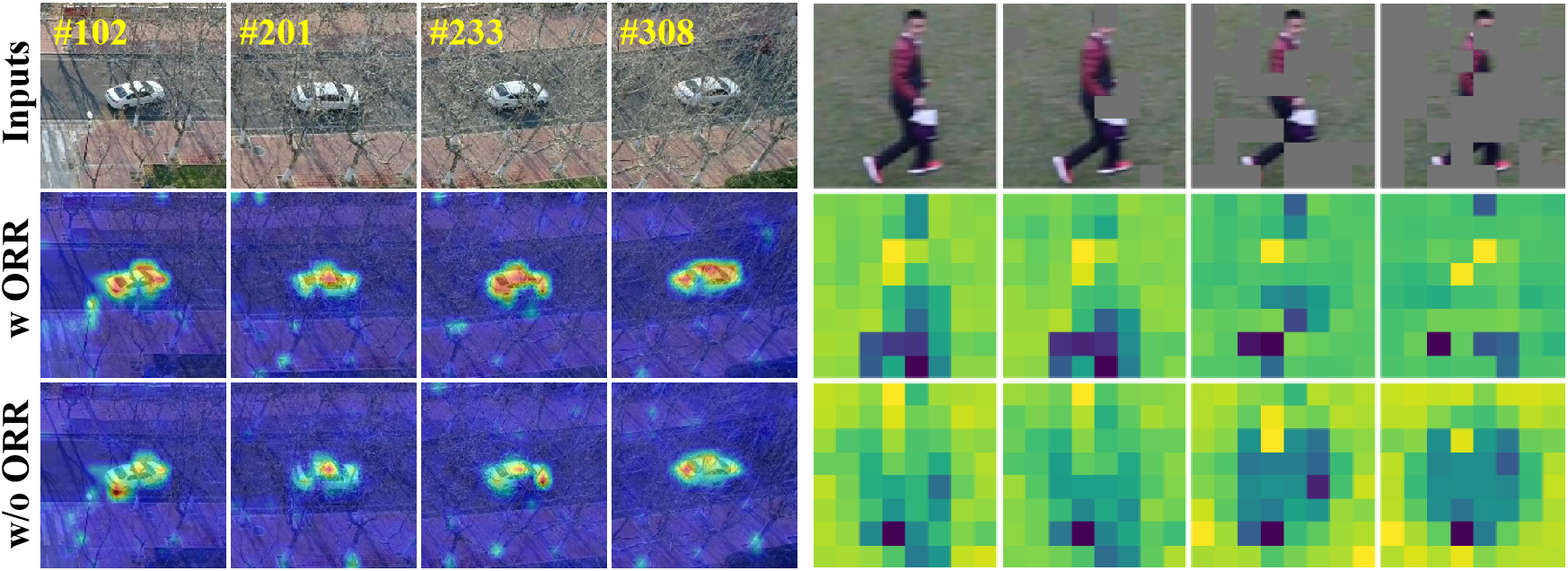}
\caption{Visualize the attention map (left) and feature map (right) of the target images. The first row displays the search and masked images with masking ratios of 0\%, 10\%, 30\%, and 70\%. The second and third rows show the attention and feature maps generated by ORTrack-DeiT, with and without ORR, respectively. } \label{fig_qualitative_res}
\vspace{-15pt}
\end{figure}

\textbf{Qualitative Results. }
Several qualitative tracking results of ORTrack-DeiT and seven SOTA UAV trackers are shown in Fig. \ref{fig_bbox_vis}. As can be seen, only our tracker successfully tracks the targets in all challenging examples, where pose variations, background clusters, and scale variations are presented. In these cases, our method performs significantly better and is more visually appealing, bolstering the effectiveness of the proposed method for UAV tracking.

Figure \ref{fig_qualitative_res} shows attention and feature maps produced by ORTrack-DeiT, with and without occlusion-robust enhancement. We observe that ORTrack-DeiT with ORR maintains a clearer focus on the targets and exhibits more consistent feature maps across masking ratios. These results support the effectiveness of our ORR component.

\section{Conclusion}

In view of the common challenges posed by target occlusion in UAV tracking, in this work, we proposed to learn Occlusion-Robust Representation (ORR) by imposing an invariance of feature representation of the target with respect to random masking modeled by a spatial Cox process. Moreover, we propose an Adaptive Feature-Based Knowledge Distillation (AFKD) to enhance efficiency. Our approach is notably straightforward and can be easily integrated into other tracking frameworks. Extensive experiments across multiple UAV tracking benchmarks validate the effectiveness of our method, demonstrating that our ORTrack-DeiT achieves SOTA performance.

\textbf{Acknowledgments. }This work was funded by the Guangxi Natural Science Foundation (Grant No. 2024GXNSFAA010484), and the National Natural Science Foundation of China (Nos. 62466013, 62206123).

\clearpage

\bibliographystyle{plain}
\bibliography{main}


\end{document}